\documentclass[a4paper, 10pt, conference]{ieeeconf}      

\IEEEoverridecommandlockouts                              

\usepackage{amsmath} 
\usepackage{amssymb}  
\usepackage{booktabs}
\usepackage{multirow}

\usepackage[inkscapepath=.]{svg}

\usepackage{atbegshi}
\usepackage{picture}

\newcommand{\myheader}{%
  Accepted to 2027 IEEE/SICE International Symposium on System Integration (SII).%
}

\newcommand{\mycopyrightnotice}{%
  \begin{minipage}{14cm}
    \scriptsize
    \copyright\ 2027 IEEE. Personal use of this material is permitted.
    Permission from IEEE must be obtained for all other uses, in any current
    or future media, including reprinting/republishing this material for
    advertising or promotional purposes, creating new collective works, for
    resale or redistribution to servers or lists, or reuse of any copyrighted
    component of this work in other works.
  \end{minipage}%
}

\title{\LARGE \bf
DRaGAN: Dynamic Graph Attention Network for Multi-Radar Point Cloud Encoding in Automotive Radar Perception
}

\author{Christof Leitgeb$^{1,2*}$, Thomas Puchleitner$^{1}$, Max Peter Ronecker$^{2}$ and Daniel Watzenig$^{2,3}$
\thanks{*Corresponding author {\tt\small christof.leitgeb@infineon.com}}
\thanks{$^{1}$Infineon Technologies AG, Austria}%
\thanks{$^{2}$Graz University of Technology, Austria}%
\thanks{$^{3}$Virtual Vehicle Research GmbH, Austria}%
}

\begin{document}

\newif\iffirstpage
\firstpagetrue

\AtBeginShipout{%
  \AtBeginShipoutUpperLeft{%
    \unitlength=1pt
    \begin{picture}(0,0)
      \put(52,-20){\parbox{\textwidth}{\scriptsize\myheader}}
    \end{picture}%
  }%
  \iffirstpage
    \AtBeginShipoutUpperLeft{%
      \unitlength=1pt
      \begin{picture}(0,0)
        \put(52,0){%
          \raisebox{-\dimexpr\paperheight-35pt\relax}[0pt][0pt]{%
            \mycopyrightnotice
          }%
        }%
      \end{picture}%
    }%
    \global\firstpagefalse
  \fi
}
\maketitle
\thispagestyle{empty}
\pagestyle{empty}

\begin{abstract}

Radar represents a key modality for weather-robust perception. Unlike optical sensors such as camera and LiDAR, radar can penetrate fog, rain and snow due to its longer wavelength, and recent hardware advances have steadily improved its spatial resolution, thus making it increasingly relevant for autonomous systems. While processing raw radar data requires substantial memory computation, most existing work has focused on preprocessed point-cloud data similar to LiDAR. However, the sparse radar points and sensor-relative anisotropy make LiDAR-oriented encoders suboptimal. To address this problem, we present DRaGAN, a dynamic graph attention network for multi-radar point clouds.  DRaGAN introduces a sensor-relative Fourier encoding of angle and range to capture fixed-aperture radar characteristics, gated short- and long-range hub nodes along each sensor’s line of sight for efficient local–global message passing, and attention-guided dynamic rewiring that prunes weak edges and adds a bounded number of latent connections restricted to two-hop neighborhoods. This reduces the number of processed edges in deeper layers while preserving representation quality. On TruckScenes, DRaGAN improves test NDS by 10.10 points over the evaluated RadarGNN baseline and reduces the reported model FLOP count by 63.7\%. This trade-off includes an increase in model latency from 40 to 53.16 ms and in trainable parameters from 1.2M to 1.7M on the evaluated implementations. Code and pretrained models are available under https://github.com/chr-is-tof/DRaGAN.
\end{abstract}

\section{INTRODUCTION}
Radar provides a complementary sensing modality for autonomous vehicles and robots, particularly in adverse weather. Due to its longer operating wavelength, radar is generally less affected by atmospheric scattering than camera or LiDAR, whose visibility, range, and measurement quality heavily degrade in fog, rain, and snow~\cite{feng_review_weather_2025}. Radar sensors are low-cost and can be mounted behind RF-transparent body panels, offering protection from direct contamination. Radar point clouds are considerably sparser than LiDAR, thereby reducing downstream data volume and enabling faster processing with lightweight models. Although radar lacks high-resolution geometric detail, it directly measures radial (line-of-sight) Doppler velocity, a strong cue for identifying and characterizing dynamic objects~\cite{yao2025exploring}.

Recent approaches have demonstrated the capability of radar-only perception by relying on raw FFT processed recordings~\cite{paek_kradar_2022, leitgeb_rade-net_2025, leitgeb_dinorade_2026}, but large data and model sizes limit real-time system integration~\cite{yao2025exploring}. Thus, a sparser data representation in the form of point-clouds is preferred. Notably, point-cloud processing techniques adapted from the LiDAR domain~\cite{Yin_2021_centerpoint} introduce significant discretization errors and computational overhead when applied to sparse points distributed over a large area. Therefore, Graph Neural Networks (GNNs) have shown strong potential for modeling non-Euclidean radar data with high accuracy~\cite{li_2024_surveygnnpointcloud}. Several methods have constructed graphs directly on radar points and applied GNN-based feature extraction for downstream tasks like object detection and segmentation~\cite{fent_radargnn_2023, svenningsson_2021_radarpointgnn, guo_2026_irgnn}. However, existing work tends to treat radar points similar to LiDAR recordings and does not consider unique recording properties of fixed-aperture radar from multiple sensors. Additionally, computationally expensive graph edge connections are currently generated statically and persist between irrelevant node pairs, which increases arithmetic load in deeper network layers. We address this research gap.

Specifically, we introduce DRaGAN, a dynamic graph attention network tailored to point clouds acquired by multiple fixed-aperture radars distributed around a vehicle. DRaGAN targets to reduce the arithmetic work of graph message passing while preserving local and cross-sensor information exchange. Unlike rotating LiDAR, fixed-aperture radar measurements exhibit viewing-angle-dependent variations in return intensity, noise, localization uncertainty, and Doppler characteristics relative to the sensor boresight~\cite{burza_radarallignment_2024}. To represent these spatially varying measurement properties, we employ a sensor-relative Fourier encoding of each point’s range and off-boresight angle, complemented by learned sensor-specific embeddings. Local message passing is performed on a graph constructed from spatial and Doppler proximity. Furthermore, attention-guided graph refinement reduces the number of node-node edges processed in deeper layers by pruning weak connections and introducing a bounded number of latent-space connections. We evaluate both operation count and realized latency because reduced arithmetic complexity does not necessarily imply faster execution for irregular sparse workloads. To enhance global context aggregation, we introduce sensor-specific short-range and long-range hub nodes that facilitate region-aware information exchange across the graph. We evaluate DRaGAN on the large-scale TruckScenes dataset~\cite{truckscenes2024}, which contains dense radar measurements from six vehicle-mounted sensors with sensing ranges of up to 200 m. 
The main contributions of this work are:
\begin{itemize} \item A sensor-relative Fourier encoding that represents the range- and viewing-angle-dependent characteristics of distributed fixed-aperture radars. \item Gated short- and long-range sensor hubs that enable controlled local-to-global and cross-sensor information exchange. \item An attention-guided rewiring strategy with bounded two-hop candidate generation that reduces node-node edge processing in deeper layers. \item A joint evaluation of detection performance, operation count, graph size, and realized GPU latency, exposing the trade-off between arithmetic complexity and sparse-operator efficiency. \end{itemize}

\section{Related Work}
\subsection{Automotive Datasets}
The growing interest in radar-based perception has led to the inclusion of high-quality radar data alongside LiDAR and camera imagery in several large-scale automotive datasets. Due to memory constraints, most datasets provide radar data exclusively as sparse point clouds~\cite{nuscenes_caesar_2020, palaffy_vod_2022, truckscenes2024}, with notable exceptions such as RaDIal~\cite{rebut_radial_2022} and RaDICal~\cite{lim_radical_2021}, which provide raw ADC time-series data, and K-Radar~\cite{paek_kradar_2022}, which offers FFT-processed tensors.

Radar data quality also varies substantially across datasets. NuScenes~\cite{nuscenes_caesar_2020}, originally designed with a focus on camera and LiDAR, provides only sparse 2D + Doppler radar points from five surrounding sensors. In contrast, View-of-Delft~\cite{palaffy_vod_2022} and TJ4DRadSet~\cite{zheng_tj4dradset_2022} offer denser 3D + Doppler points, but from a single front-facing sensor only. This increased data density has enabled current radar-only detection methods to achieve competitive performance. However, such single-sensor setups fail to reflect real-world automotive configurations, where radar sensors are distributed around the vehicle to capture the complete surrounding scene.

To evaluate multi-sensor radar point-cloud perception under realistic operating conditions, we use the TruckScenes dataset~\cite{truckscenes2024}, which provides dense 3D + Doppler radar measurements from six sensors distributed around a truck and covers diverse long-range driving scenarios up to 200 m, including adverse weather conditions.

\subsection{Automotive Radar Perception}
Automotive 77 GHz radar sensors directly measure range and Doppler velocity with high precision while estimating the angle of arrival. By applying Fast Fourier Transforms (FFT) to the mixed transmit and receive signals, the sensor generates a multi-dimensional spectrum whose peaks encode target position and velocity. Peak detection is then used to produce a radar point cloud~\cite{srivastav_radarsfa_2023}. As a result, point characteristics depend on their location relative to the radar boresight~\cite{burza_radarallignment_2024}. While this representation reduces downstream memory requirements, it discards substantial information contained in the original radar spectrum~\cite{leitgeb_rade-net_2025}.

Recent ML-based radar perception methods have demonstrated a substantial performance improvement by learning from denser raw data representations, such as FFT spectra~\cite{leitgeb_dinorade_2026, leitgeb_rade-net_2025} or reduced radar cube~\cite{paek_kradar_2022}. In contrast, radar point clouds have typically been processed using voxel- or pillar-based encoding methods adapted from the LiDAR domain~\cite{yin_centerpoint_2021}. However, the resulting feature maps are often too sparse for radar-only perception, necessitating fusion with complementary modalities such as LiDAR or camera~\cite{wolters2024sparc, wolters_hydra_2025}. SpaRC~\cite{wolters2024sparc} proposes a sparse radar-camera fusion transformer that combines image features and transformer-encoded radar point clouds through multi-stage cross-attention, using range-adaptive radar refinement and query-based object decoding to improve 3D object detection performance. HyDRa~\cite{wolters_hydra_2025} improves radar-camera fusion by jointly leveraging radar and image features for depth-aware BEV representation learning, achieving state-of-the-art performance in 3D object detection and semantic occupancy prediction.

\subsection{Graph-based Point Cloud Encoding}
Point clouds have become a popular representation for 3D sensory data such as radar and LiDAR. However, the non-euclidean irregularity of point clouds presents numerous challenges for traditional deep learning frameworks~\cite{li_2024_surveygnnpointcloud}. Graph neural networks (GNNs) have emerged as effective methods for point-cloud processing because they naturally handle unordered, irregularly distributed detections while capturing local geometric relationships through graph-based message passing. Point-GNN~\cite{shi2020point} demonstrated that graph neural networks can operate directly on irregular LiDAR point clouds by modeling individual points as graph nodes and exchanging information through graph convolutions. The ability of GNNs to capture local geometric relationships in non-Euclidean point-cloud data motivated their adoption for naturally sparse automotive radar measurements, where graphs can be constructed directly on radar detections while additionally exploiting radar-specific attributes such as Doppler velocity~\cite{fent_radargnn_2023,svenningsson_2021_radarpointgnn,guo_2026_irgnn}. Radar-PointGNN~\cite{svenningsson_2021_radarpointgnn} constructs graph nodes directly from radar detections and iteratively propagates information through graph convolutions to learn contextual object representations from translation invariant features. RadarGNN~\cite{fent_radargnn_2023} formulated a more general message passing neural network (MPNN) layer than graph convolution in\cite{svenningsson_2021_radarpointgnn} to consider edge features within the update process. The authors observed significant performance improvement for translation invariant features where global node coordinates are replaced by relative distance within the respective connecting edges while joint translation and rotation invariance showed no improvement. IRGNN~\cite{guo_2026_irgnn} employs invariant graph representations along with an enhanced MPNN backbone that incorporates residual connections and virtual node layers for improved global context aggregation. Notably, existing radar-based GNN approaches rely on statically constructed graphs whose connectivity remains fixed throughout processing. Our work addresses this limitation by dynamically removing irrelevant edges and establishing new connections as node representations evolve.

\begin{figure*}
    \centering
    \includesvg[width=1\linewidth]{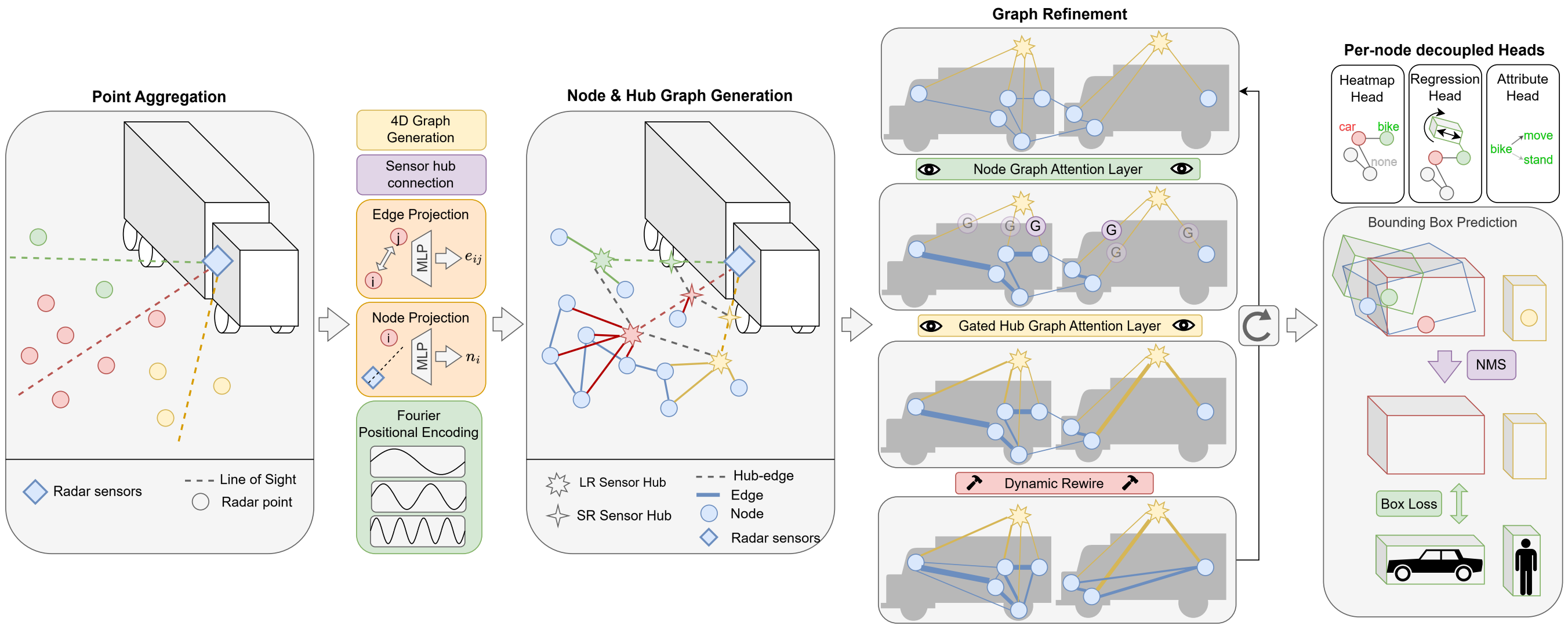}
    \caption{DRaGAN methodology overview.}
    \label{fig:architecture_overview}
\end{figure*}
\section{Methodology}
Fig~\ref{fig:architecture_overview} shows an overview of our proposed method. In contrast to existing graph-based approaches, we specifically tailor the encoding towards the physical properties of radar points coming from different sensor lines of sight (LoS) while maintaining a compact, dynamically adjusted graph size with local and global information exchange. First, a set of radar points $p_i^s \in \mathbb{R}^{C_p}$ is collected from each sensor, where $C_p$ consists of position $(x,y,z)$, radial Doppler velocity $v_d$, radar cross section $\sigma$ and sensor index s. Second, an initial graph $\mathcal{G}$ is constructed based on k nearest neighbors (kNN) both in the spatial and Doppler domain, so that connectivity reflects rigid body motion coherence rather than spatial proximity alone. Since information exchange within $\mathcal{G}$ is limited to local neighborhoods, we adapt the concept of virtual nodes~\cite{gilmer_neural_2017} for stronger global context aggregation. Specifically, we extend $\mathcal{G}$ by a set of short-range and long-range hubs located along the LoS of each sensor and additional edge connections to all points located inside the respective radar range. After graph construction, node features $N_n$ and hub features $H_h$ as well as connecting edge features $E_{nn}$, $E_{nh}$ and $E_{hh}$ are projected from the radar node and edge attributes. Next, we add a Fourier positional encoding which models the relative position of each radar point within the field of view of the respective sensor. Furthermore, the graph is iteratively refined by graph attention and dynamic rewiring layers. The GATv2Conv\cite{brody2022how} based graph attention layer computes attention weights for all edge connections and performs weighted message passing between relevant nodes. To control graph complexity, each subsequent dynamic rewiring layer prunes weak node-node connections based on low attention scores and rewires previously unconnected nodes based on a latent similarity score of two-hop neighbors. After the graph refinement, three decoupled prediction heads estimate bounding box parameters, class labels and attributes for all node proposals which are then reduced and filtered by confidence to get the final set of object proposals.

\subsection{Radar-aware Graph Construction}
The overall graph structure $\mathcal{G}$ consists of node attributes 
\begin{equation}
    N_i = (x_i, y_i, z_i, v_i^x, v_i^y, v_i^z, \sigma_i, s_i)
\end{equation}
and hub attributes 
\begin{equation}
    H_k = (x_k, y_k, z_k, v_{ego}^x, v_{ego}^y, v_{ego}^z, \alpha d, s_k)
\end{equation}
where $(x_k, y_k, z_k)$ represents the position along the line of sight of $s_k$ with distance $d=d_s$ for short-range hubs and $d=d_l$ for long-range hubs and $\alpha$ is a scaling factor. Additionally, the radial Doppler velocity $v_d$ is decomposed into three spatial aligned components $(v_i^x, v_i^y, v_i^z)$. The corresponding edge attributes 
\begin{equation}
    E=(\Delta x, \Delta y, \Delta z,\Delta v^x,\Delta v^y,\Delta v^z,\Delta\sigma)
\end{equation}
are defined (I) for nodes and their respective k nearest neighbors based on spatial and Doppler distance; (II) for all $\binom{12}{2}=66$ unordered hub pairs, represented by 132 directed hub-hub pairs; (III) for node-hub pairs where $s_i=s_k$ and the node falls in the respective long- or short-range region. After graph construction, all attributes are mapped into learned 64-dimensional latent representations $N'$, $H'$ and $E'$.

\subsection{Sensor-Relative Fourier Positional Encoding}
Since the angular resolution and reflection characteristics of a fixed-aperture radar vary with a point's position within the sensor's field of view~\cite{burza_radarallignment_2024}, each node is further characterized by its off-boresight angle $\beta_i$ and logarithmic range $\mu_i = \log(1{+}r_i)$, both defined relative to the originating sensor rather than the ego frame. These quantities are encoded via Fourier features at $F$ increasing frequencies $f_k$, using the shorthand $\mathbf{c}(\theta) = (\cos\theta, \sin\theta)$,
\begin{equation}
    \gamma(p_i) = \big[\, \mathbf{c}(f_k \beta_i),\ \mathbf{c}(f_k \mu_i) \,\big]_{k=0}^{F-1},
\end{equation}
yielding a smooth representation of angle alongside a compact representation of range. To explicitly convey the increasing angular and cross-range uncertainty of returns observed near the periphery of the sensor's field of view, two additional scalars proportional to $1/\cos\beta_i$ are appended, rather than relying on the network to infer this behavior from $\gamma(p_i)$ alone. The resulting vector is linearly projected and combined with the node embedding along with a learned per-sensor identity embedding, enabling geometric reasoning to be shared across sensors while preserving the capacity to learn sensor-specific calibration effects.

\subsection{Node and Hub Graph Attention}
The graph attention aims to compute the relevance between neighboring nodes, hubs, and node-hub pairs to support efficient information aggregation from multiple instances that carry semantic information about an object. The implementation is split into two parts with independent parameters. First, the node graph, represented by $N'$ and $E_{nn}'$, is propagated through $L$ stacked GATv2Conv~\cite{brody2022how} layers to perform inter-node message passing, where the edge count per node is usually small. Second, the hub graph, which includes $H'$, $N'$, $E_{hh}'$, and $E_{nh}'$, is passed through a separate set of $L$ GATv2Conv layers to perform inter-hub and hub-node information exchange, avoiding attention dilution that a single joint softmax would otherwise suffer from given the large disparity in node degree between hubs and ordinary nodes. Each hub receives aggregated messages weighted by attention scores from numerous nodes and from every other hub, while each node only receives one message, from its connected hub.

We observed that these hub-node messages may not be useful to all nodes and can even act as noise for some of them. To address this, each layer is equipped with a learned, channel-wise gate that conditions on both the node's current state $h_i$ and the incoming hub message $m_i$,
\begin{equation}
    g_i = \sigma\big(W_g\,[h_i \,\|\, m_i]\big), \qquad h_i' = h_i^{\text{local}} + g_i \odot m_i,
\end{equation}
where only relevant feature channels of $m_i$ are passed through. The gate weights are initialized to zero with a negative bias, so hub information is largely suppressed at the start of training and can be selectively opened up where it is beneficial.

\subsection{Dynamic Rewiring}
Since the initial constructed graph $\mathcal{G}$ was generated solely based on spatial and Doppler proximity, it does not necessarily reflect the true node relations required to exchange meaningful object-specific context as well as contain unnecessary or weakly contributing graph structures that increase computation and memory. Following the notion of recomputing graph connectivity from evolving node features across layers~\cite{wang_2019_DGCNN}, the node-node edges are rewired after every GATv2Conv~\cite{brody2022how} layer using the computed attention scores and updated latent representations. Meanwhile, the hub structure is left unchanged since including it in the candidate search would let the high in-degree of hub nodes dominate the two-hop neighborhood computation. The dynamic rewiring is performed in three steps:

\paragraph{Pruning} Attention weights for an edge $E'_{nn}(i,j)$ are averaged over all heads and the edge is retained only if it exceeds the mean attention received by its target node $N'_n(j)$. This allows the network to remove edges itself judged uninformative.

\paragraph{Candidate Generation} Potential candidate pairs to form a new edge connection are restricted to each node's two-hop neighborhood in the pruned graph, obtained via a single sparse adjacency-squaring operation. This limits the candidate set to $\mathcal{O}(N \cdot \bar{k}^2)$ pairs, where $\bar{k}$ is the average node degree, rather than the $\mathcal{O}(N^2)$ pairs a global search over the latent space would require.

\paragraph{Edge Generation} Candidates are ranked by squared distance metric in the updated latent space where only the $k'$ nearest candidates per node are kept which allows the graph to form connections between nodes the network currently considers similar. As there are no physical edge attributes available for these latent-only connections, a learned linear projection is used in place:
\begin{equation}
    E''_{nn}(i,j) = \operatorname{W_e}(N'_n(j)- N'_n(i))
\end{equation}
Finally, the updated node edges are combined with the unchanged hub edges and forwarded to the subsequent attention layer.

\subsection{Object Detection}
Following a center-based detection paradigm~\cite{Yin_2021_centerpoint}, the refined graph's node features are passed through three decoupled and task-specific detection heads which estimate bounding-box parameters, class labels and corresponding attributes. All heads are implemented as an MLP structure to operate directly on the graph in a per-node style. (I) The heatmap head proposes foreground object classes along with a confidence score at each spatial node location. (II) The regression head predicts 3D bounding box parameters $\rho_i = [dx,dy,dz,l,w,h,sin\theta,cos\theta,v_x,v_y]$, where $(dx,dy,dz)$ represent the offset from the respective spatial node location. (III) The attribute head estimates class specific attributes which match the corresponding predicted object class. After the prediction, all boxes are assembled and per-class non-maximum suppression (NMS) is applied to remove duplicate proposals before arriving at the final set of boxes. 

\subsection{Loss}
The overall loss consists of four independent weighted terms: A focal-based heatmap loss $\mathcal{L}_h$ to optimize the object proposal prediction, a Smooth L1-based regression loss $\mathcal{L}_r$ to refine bounding box parameters $\rho_i$, a cross-entropy attribute loss $\mathcal{L}_a$ and, similar to~\cite{yang_rethinking_2022}, an auxiliary Gaussian Wasserstein distance (GWD) loss $\mathcal{L}_{gwd}$.
\begin{equation}
    \mathcal{L} = \alpha_h \mathcal{L}_h + \alpha_r \mathcal{L}_r + \alpha_a \mathcal{L}_a + \alpha_{gwd} \mathcal{L}_{gwd}
\end{equation}
While $\mathcal{L}_h$ is computed over all nodes, $\mathcal{L}_r$, $\mathcal{L}_a$ and $\mathcal{L}_{gwd}$ are only computed for nodes that fall inside a ground-truth bounding box and thus consider specific parameters from the respective box. Specifically, the GWD loss models each box as a Gaussian and measures distributional distance to the ground-truth box, refining shape and orientation beyond what SmoothL1 alone captures, since SmoothL1 treats each regression channel independently and does not directly penalize orientation and shape mismatch as a geometric quantity.

\begin{table*}[htbp]
\centering
\caption{Comparison of different modality methods on the TruckScenes benchmark.}
\setlength{\tabcolsep}{8pt}
\small
\begin{tabular}{lcccccccc|c}
\toprule
Method & Modality & NDS$\uparrow$ & mAP$\uparrow$ & mATE$\downarrow$ & mASE$\downarrow$ & mAOE$\downarrow$ & mAVE$\downarrow$ & mAAE$\downarrow$ & Split \\
\midrule
CenterPoint~\cite{yin_centerpoint_2021} & L & \underline{40.96} & \underline{26.70} & \textbf{0.4089} & \textbf{0.3525} & \textbf{0.2769} & 2.7301 & \underline{0.2008} & \multirow{6}{*}{Test} \\
PETR~\cite{petr_2022_liu} & C & 12.14 & 2.22 & 1.1252 & 0.6863 & 0.6466 & 1.4987 & 0.5643 & \\
SpaRC~\cite{wolters2024sparc} & R+C & \textbf{42.04} & \textbf{33.22} & 0.7174 & \underline{0.3932} & \underline{0.3545} & \textbf{0.7841} & 0.2077 & \\
HyDRa~\cite{wolters_hydra_2025} & R+C & 22.37 & 12.78 & 0.7253 & 0.5441 & 0.7442 & 1.1804 & 0.3883 & \\
RadarGNN~\cite{fent_radargnn_2023} & R & 10.78 & 6.98 & 0.8917 & 0.8087 & 1.1323 & 8.0028 & 0.5706 & \\
DRaGAN (ours) & R & 20.88 & 8.76 & 0.7277 & 0.6388 & 0.6442 & \underline{1.0438} & 0.3391 & \\
\midrule
CenterPoint-V*~\cite{yin_centerpoint_2021} & L & \underline{35.30} & \textbf{22.60} & \textbf{0.4610} & \textbf{0.4050} & \textbf{0.4680} & 3.0280 & 0.2610 & \multirow{4}{*}{Val}\\
Far3D~\cite{jiang_far3d_2024} & C & 21.40 & 10.70 & 0.8830 & 0.5070 & 0.6710 & 1.3520 & 0.3380 & \\
SpaRC~\cite{wolters2024sparc} & R+C & \textbf{35.40} & \underline{22.50} & 0.7980 & \underline{0.4490} & \underline{0.4760} & \textbf{0.6130} & \underline{0.2480} & \\
DRaGAN (ours) & R & 23.25 & 6.60 & \underline{0.7698} & 0.5894 & 0.5617 & \underline{1.0942} & \textbf{0.0839} & \\
\bottomrule
\end{tabular}
\vspace{2pt}
{\\ \footnotesize best result in \textbf{bold} and second best result in \underline{underline}.}
\label{tab:benchmark}
\end{table*}

\section{Experiments}
To validate our method, we conduct experiments on the MAN TruckScenes~\cite{truckscenes2024} dataset, which provides dense radar point clouds from multiple sensors in different orientations which makes it an excellent fit. Our evaluation is performed according to the TruckScenes benchmark and results are compared with various multi-modal methods presented on the official test server as well as on the validation split. As primary metric we use the NuScenes Detection Score~\cite{nuscenes_caesar_2020}, which is defined by the aggregation of mean Average Precision (mAP) and five mean True Positive (mTP) metrics.
\begin{equation}
    \text{NDS} = \frac{1}{10} \left[ 5 \cdot \text{mAP} + \sum_{\text{mTP} \in \mathbb{TP}} (1 - \min(1, \text{mTP})) \right]
\end{equation}
where we additionally define the contribution from the mTP metrics in one score.
\begin{equation}
\label{eq:mtps}
    \text{mTPS} = \sum_{\text{mTP} \in \mathbb{TP}} (1 - \min(1, \text{mTP}))
\end{equation}
We compare the NDS results from the official test server and the validation split in Table~\ref{tab:benchmark} and include mAP and all five error metrics to discuss model specific capabilities. For a detailed description of the metrics we refer to~\cite{nuscenes_caesar_2020}.

\subsection{Training}
We train the model on a NVIDIA L40S GPU for 20 epochs using a batch size of 8. Optimization is performed using AdamW with an initial learning rate of $5 \times 10^{-4}$ and a weight decay of $10^{-4}$. The learning rate is adjusted after each epoch using a cosine annealing schedule with a minimum learning rate of $10^{-5}$. The model contains $1.7$ million trainable parameters.

\subsection{Computational Complexity and Runtime}
We evaluate real time performance and estimated computing requirements for edge platforms. Measurements are performed on the NVIDIA L40S with FP32 precision. We compare inference time, floating point operations (FLOPs), multiply-accumulate operations (MACs) and parameter counts with RadarGNN~\cite{fent_radargnn_2023}. Table~\ref{tab:realtimeperformance} shows that our method requires higher inference time (32.9\%) than RadarGNN while reducing FLOP and MAC counts by 63.7\%. The reduction in reported FLOPs and MACs does not translate into lower latency, where DRaGAN takes 53.16 ms versus 40 ms for RadarGNN and has 1.7M compared to 1.2M trainable parameters. Irregular operations such as dynamic rewiring and sparse message passing may contribute to the higher GPU latency. However, we do not isolate their contributions in this study.

\begin{table}[htbp]
\centering
\caption{Model-forward complexity and runtime at batch size 1 and FP32 precision.}
\begin{tabular}{lcc}
\hline
\textbf{Metric} & \textbf{RadarGNN~\cite{fent_radargnn_2023}} & \textbf{DRaGAN (ours)} \\ \hline
Inference Time  & $40 \pm 0.7$ ms    &  $53.16 \pm 0.18$ ms \\
FLOPs           & $63 \times 10^{9}$ & $22.9 \times 10^{9}$     \\
MACs            & $32 \times 10^{9}$ & $11.5 \times 10^{9}$     \\
Parameters      & $1.2 \times 10^{6}$ & $1.7 \times 10^{6}$    \\ \hline
\end{tabular}
\label{tab:realtimeperformance}
\end{table}

\subsection{Results}
The quantitative results in Table~\ref{tab:benchmark} show that DRaGAN improves all reported detection metrics over the evaluated RadarGNN~\cite{fent_radargnn_2023} on the test split, increasing NDS from 10.78 to 20.88. The accuracy gain is accompanied by higher model-forward latency and a larger parameter count (Table~\ref{tab:realtimeperformance}). Whether this trade-off is preferable depends on the deployment constraints. DRaGAN’s mAAE increases from 0.0839 on validation to 0.3391 on test. This gap may partly reflect differences in the difficulty and attribute composition of matched detections across the splits, since mAAE is evaluated on matched detections rather than all ground-truth objects. Additionally we outperform all camera-only approaches~\cite{jiang_far3d_2024, petr_2022_liu} while getting close to HyDRa~\cite{wolters_hydra_2025} radar-camera fusion. The evaluation demonstrates that our method achieves superior learning of object properties and motion attributes once objects are detected in the scene which suggests strong information exchange among nodes and effective use of radar specific attributes. Due to the special focus on including Doppler velocity in the graph construction as well as in both, node and edge features, DRaGAN reports the second lowest mean average velocity error (mAVE) on the test and validation set. We can observe that radar-only methods still show a lower mAP compared to other modalities, which is primarily attributed to complex scenes where sparsity still poses a challenge in separation of adjacent objects. However, the results show a promising behavior of radar-only sparse point-cloud perception on a challenging dataset.

\subsection{Ablation Study} Table~\ref{tab:ablation_study} evaluates the contribution of the proposed architectural components by removing each component individually from the full DRaGAN model. Results were averaged over three runs to mitigate stochastic variance and reduce the likelihood of fortuitous outcomes. We compare the metrics NDS and mAP described in~\cite{nuscenes_caesar_2020} as well as the mTPS contribution from Equation~\ref{eq:mtps}.

\paragraph{Fourier Positional Encoding}
Removing the Fourier encoding primarily reduces mTPS, indicating that sensor-relative geometry and off-boresight information improves the quality of estimated object properties after detection, rather than the number of correctly detected objects.

\paragraph{Dynamic Graph Rewiring}
A static graph degrades both mAP and mTPS and increases the inference time by 9.45 ms by retaining more edges in deeper layers. Averaged over the validation set, dynamic rewiring reduces the number of local edges from 364,662 in the first layer to 80,969 in the fourth layer, corresponding to a reduction of 77.8\%. Since the 46,637 hub edges remain fixed, the total edge count decreases from 411,299 to 127,606, a reduction of 69.0\% which results in faster processing (9.45 ms).

\paragraph{Sensor Hubs}
Removing the sensor hubs causes a substantial mTPS reduction while mAP is significantly lower affected. This ablation tests the hub pathway, not the gate in isolation. The results are consistent with the hypothesis that hub-mediated context primarily improves bounding-box, motion, and attribute estimation for detected objects. It further supports the role of the hubs in exchanging contextual information across spatially separated radar regions.

\paragraph{Graph Attention}
Removing graph attention causes the largest degradation in both mAP and mTPS, demonstrating that graph attention has the largest measured contribution among the evaluated components.

\paragraph{Ego-Frame Coordinates}
Removing ego-frame coordinates primarily reduces mTPS, suggesting that explicit
scene-level context benefits the integration of measurements from distributed
radars. This differs from RadarGNN~\cite{fent_radargnn_2023}, where
translation-invariant features were more effective.

\begin{table}[htbp] \centering \caption{Ablation study on the validation split.} \begin{tabular}{ccccccc} \hline Metric & (a) & (b) & (c) & (d) & (e) & DRaGAN \\ \hline \hline 
NDS$\uparrow$ & 19.49 & 20.07 & 15.23 & 10.59 &  17.63 & 23.25 \\ 
mAP$\uparrow$ & 6.23 & 5.09 & 5.9 & 2.15 & 6.05 & 6.60 \\ 
mTPS$\uparrow$ & 163.75 & 175.25 & 122.80 & 95.15 & 146.05 & 199.50 \\ 
\hline \end{tabular} \label{tab:ablation_study} \end{table}

\section{Conclusion}
We presented DRaGAN, a multi-radar-aware graph neural network that combines sensor-relative spatial encoding, gated sensor hubs, graph attention, and dynamic graph rewiring for radar-only 3D perception in autonomous driving. On the official TruckScenes benchmark, DRaGAN improves NDS by 10.10 points over the GNN-based radar-only baseline while reducing the estimated FLOP count by 63.7\%. This improvement demonstrates the value of jointly modeling sensor-specific viewing geometry, Doppler-aware local relationships, and cross-sensor context. The lower arithmetic complexity currently comes with a 13.16 ms (32.9\%) increase in model-forward latency on the tested GPU, potentially reflecting the cost of irregular sparse graph operations, whose individual contributions were not measured. These results support dynamic graph refinement as a way to reduce reported arithmetic operation count while improving detection performance relative to the evaluated baseline. The higher measured latency and parameter count remain limitations for latency- and memory-constrained deployment, which will be covered in future work.
\section{Acknowledgment}
This work is supported by Infineon Technologies Austria AG and the European Union through the Horizon Europe programme (Grant Agreement project 101092834).
\bibliographystyle{IEEEtran}  
\bibliography{references}

\end{document}